\documentclass[runningheads]{llncs}

\usepackage{eccv}

\usepackage{eccvabbrv}

\usepackage{graphicx}
\usepackage{amsmath}
\usepackage{amssymb}
\usepackage{booktabs}

\usepackage[accsupp]{axessibility}  

\usepackage{hyperref}

\usepackage{orcidlink}

\begin{document}

\title{BRo-JEPA: Learning Modular Transformations in Latent Space} 


\author{Divyansh Jha\inst{1,2} \and
Yuanfang Xie\inst{1} \and
Brennen Yu\inst{1} \and
Varan Mehra\inst{1}}

\authorrunning{D.~Jha et al.}

\institute{Georgia Institute of Technology, Atlanta, GA 30332, USA \and
New York
University, Grossman School of Medicine, New York, NY, USA, 10016}

\maketitle

\begin{abstract}
Can neural networks learn algebraic rules from visual inputs, or do they merely fit observed patterns? We study this question using MNIST (or EMNIST letters) as states and modular arithmetic operations as actions in a JEPA-style world model. Standard supervised and JEPA baselines with operation embeddings achieve high accuracy on seen operations but fail to extrapolate reliably to unseen operations. 
We propose BRo-JEPA, a world model with a block-rotation predictor that represents arithmetic operations as rotations, resulting in the cyclic structure of modular arithmetic in latent space. By applying actions as rotations, the BRo-JEPA predictor learns the rotation angles to align the latent representations with the underlying modular structure which enables strict zero-shot operation generalization. 
While our best block-rotation supervised baseline reaches only 54.54\%  zero-shot accuracy on MNIST and 25.13\% on EMNIST, BRo-JEPA with a ResNet-18 encoder achieves 99.44\% and 94.35\% respectively, despite being trained only on the primitive operations ±1. 
Our results suggest that world models can learn algebraic rules when the latent transformations encode the underlying modular structure. 
Code is available \href{https://github.com/DL-World-Models/brojepa}{here}. 

\end{abstract}

\section{Introduction}

 Mathematical and logical reasoning are regarded as crucial steps in building intelligent machines\cite{testolin2024can}. Current deep learning models have shown notable success in mathematical problems ranging from word problem solving to automated theorem proving. However, it remains unclear whether these models truly possess numeric reasoning capabilities, as they often fail to extrapolate beyond the numerical ranges or operations seen during training \cite{trask2018neural, testolin2024can}. A central challenge for deep learning is this out-of-distribution compositional generalization, where generic unstructured neural networks often struggle to extrapolate learned rules to unseen combinations \cite{Qiuetal2022}. While large-scale training can sometimes induce compositional generalization if the task space is sufficiently covered \cite{Redhardtetal2025}, targeted architectural biases or pretraining paradigms are usually required to embed the abstract representations necessary to mimic human-like understanding \cite{Itoetal2022}. 
 
 In deep learning, arithmetic is typically framed as a supervised prediction problem: given an input and an operation, the model predicts the resulting output. To better support mathematical operations, special neural arithmetic modules such as NALU and NAU introduce inductive biases for operations including addition, subtraction, multiplication, and division \cite{trask2018neural, madsen2020neural}. Although these special structures improve arithmetic learning through built-in structural priors, they still suffer from robustness and generalization \cite{mistry2022primer}.
 
 Recently, world models have demonstrated great potential in advanced machine intelligence \cite{assran2025vjepa, nature_world_models_2026} . By learning abstract semantic representation and modeling dynamics directly in the compact latent space, world models provide an efficient architecture for machine understanding, prediction, and planning \cite{assran2025vjepa}. Such learning and controlling in the latent space offers a natural framework for mathematical reasoning where mathematical rules can be represented as a structural transformation within the semantically meaningful latent space. 
 


In this work, we build on the Joint Embedding Predictive Architecture (JEPA) \cite{assran2023self}. Unlike traditional approaches that predict raw outputs, JEPA focuses on modeling transformations in the latent space. While JEPA-based approaches have shown strong performance in continuous settings involving deterministic spatial or temporal transformations, their effect on symbolic, axiomatic transformations remains largely unexplored.  




We used JEPA training framework on the MNIST and EMNIST-Letters datasets \cite{6296535,cohen2017emnist}. The model receives an image of a number / a letter (the context image), along with a mathematical operation (the action, such as "+3"), and predicts the representation of the resulting number / letter (the target). A context encoder extracts the representation from the context images. A predictor then takes this representation, applies the action, and predicts the representation of the correct answer. The model is trained by making the predicted representations and target representations as close as possible. 

In addition to the standard neural network predictor commonly used in JEPA models, we also introduced a novel block-rotation predictor that enables strong zero-shot generalization in modular arithmetic tasks. We refer to this framework as BRo-JEPA. By incorporating inductive biases aligned with the cyclic structure of modulo arithmetic, BRo-JEPA organizes latent representations into a structured geometric manifold, enabling efficient learning with a relatively simple predictor architecture. More broadly, our results suggest that imposing appropriate transformation structure in latent transition models may be a promising direction for representation learning in other domains with modular, cyclic, or group-structured dynamics.



Our contributions are:
(1) We introduce MNIST / EMNIST modular arithmetic as a controlled testbed for studying rule learning in latent world models.
(2) We propose BRo-JEPA, a JEPA-style latent dynamics model with an operation-conditioned block-rotation predictor.
(3) We benchmark across all baseline models: supervised baselines with or without block rotation, JEPA models, and JEPA models with additive operation embeddings; all fit seen operations but fail strict zero-shot extrapolation.
(4) We demonstrate strong zero-shot for both fixed- and learnable-angle BRo-JEPA, and show that learnable rotations recover frequencies compatible with the underlying modulo structure from primitive operations alone. 






\section{Approach}

\begin{figure}[t]
\centering
\includegraphics[width=0.65\linewidth]{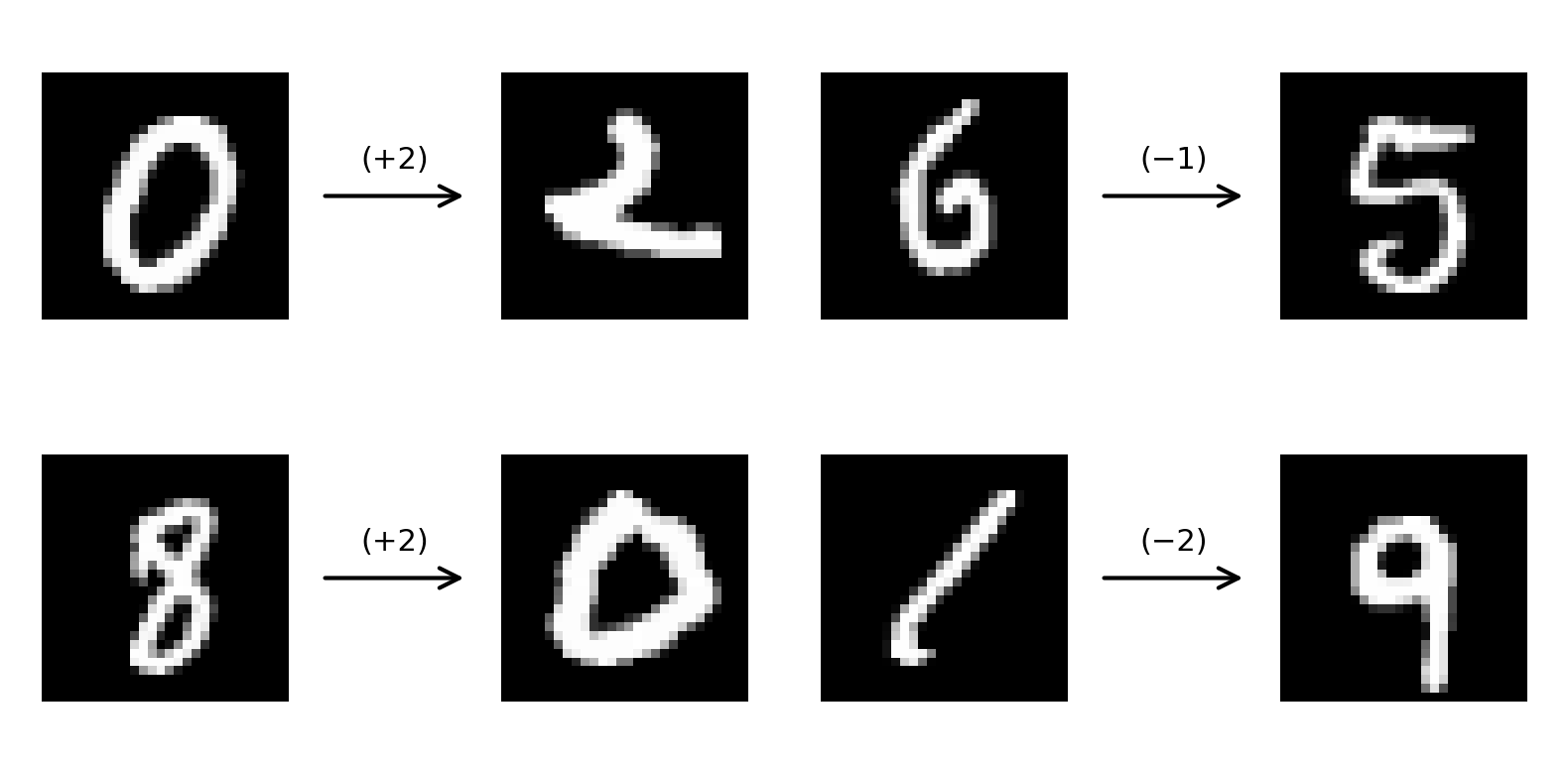}
   \caption{Qualitative examples of visual arithmetic. Given a context digit $x$ and operation $k$, the model predicts the target digit $y=(x+k)\bmod 10$.}
\label{fig:MNIST_arithmetic_examples}
\end{figure}

We utilize JEPA \cite{assran2023self} to model arithmetic over images. The model predicts the latent representation of a target digit given a context digit image and an arithmetic operation (Fig.~\ref{fig:MNIST_arithmetic_examples}). We expect this to work because arithmetic operations define consistent transformations between digit states, which can be modeled more naturally in latent space. We propose a novel use of RoPE-like \cite{su2021roformer} rotation dynamics in a world-modeling setup.

\subsection{Baseline JEPA Model}
\begin{figure}[h!]
\centering
\includegraphics[width=\linewidth]{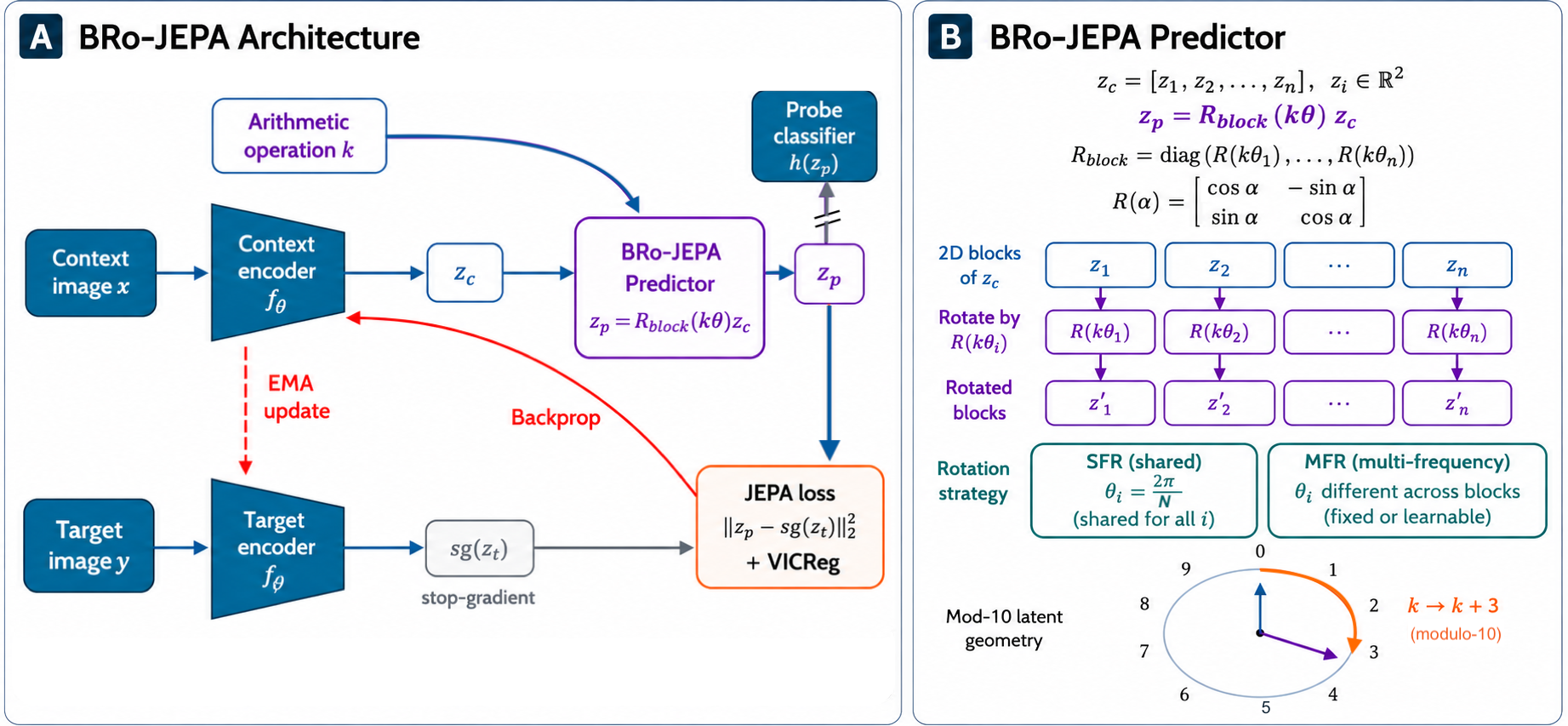}
   \caption{
Overview of BRo-JEPA.
(A) The context encoder maps image $x$ to $z_c$, and the BRo-JEPA predictor applies operation $k$ to produce the predicted target latent $z_p$, trained with a JEPA/VICReg loss against $\mathrm{sg}(z_t)$.
(B) 
BRo-JEPA splits $z_c$ (dimension of $2n$) into $n$ 2D blocks and applies operation-conditioned rotations $R(k\theta_i)$, yielding $z_p = R_{\mathrm{block}}(k\theta)z_c$ and encoding modular arithmetic as cyclic latent dynamics. In SFR, $N$ is the number of classes in the dataset, which can be learned. 
}
\label{fig:fig1}
\label{fig:our_jepa_figure}
\end{figure}

We adopt the standard JEPA architecture (Fig. \ref{fig:fig1}A) \cite{assran2023self}. The context encoder maps the context image to a latent representation, which is concatenated with the operation embedding. The target encoder maps the target image to a latent representation and is updated as an exponential moving average (EMA) of the context encoder , a key stabilizing mechanism for preventing representation collapse \cite{Grill2020}. An MLP predictor takes the context–operation latent as input and predicts the target latent. The L2 loss is computed between the predicted latent and the target latent. A classifier is attached to the predicted latent, serving as an online probe to quantify the semantic quality of learned representations. The classifier is detached in all of our JEPA experiments. Its cross-entropy loss does not backpropagate through the predictor or encoders. 
Thus, the classifier provides no supervised signal to JEPA representation learning.

\subsection{Intermediate Compositional Baselines}

Before introducing BRo-JEPA, we first tested whether compositionality could be induced through the operation representation or through an auxiliary training objective while keeping the predictor unconstrained. These variants are not our main proposed architecture; rather, they serve as diagnostic baselines that help identify why a stronger structural bias is needed.

\textbf{Additive Compositional Action Embedding}
We introduced an additive compositional embedding for actions, where each operation ($op(k)$) is represented as a scalar multiple of a learned base vector ($\mathbf{v}$) of the primitive operations (+1 or -1) i.e. $op(k) = k\cdot \mathbf{v}$.  This parameterization enforces an arithmetic structure by construction, ensuring that relationships such as $op(+3)=3 \cdot op(+1)$ hold exactly. This design encodes linear compositionality directly in the embedding space. A list of other embedding methods that we tried is present in Appendix \ref{sec:op_embeddings}.

\textbf{Composition Consistency Loss}
To complement the embedding, we introduced a composition consistency loss that explicitly enforces compositional behavior in the predictor. The key idea is that applying the composed operation in one step should match applying its primitive operation sequentially in the predicted latent space. Formally, for the predictor $g(\phi) = g(z, \mathbf{v})$, we enforce $g(z, k \cdot \mathbf{v}) \approx g(\underbrace{g(g(\dots g(z, \mathbf{v}) \dots), \mathbf{v})}_{k \text{ times}})$, where $z$ is the context latent representation and $\mathbf{v}$ is the primitive operations (+1/-1). Because sequentially applying primitive operation has shown great robustness in our rollout evaluation, we can use the sequential rollout predicted latent as the target for our composition consistency loss.  Let's take +2 for an example, where the sequential rollout predicted latent $z_{seq} = g(g(z, +1),+1)$, and the direct one-step predicted latent $z_{direct} = g(z,+2)$. The composition consistency loss is computed as $||z_{direct} - z_{seq}||^2$ (details in Appx. \ref{sec:composition_consistency_loss}). This auxiliary objective is not a strict zero-shot setting because it evaluates the predictor at composed operation values during training. We include it as an intermediate baseline to test whether enforcing compositional consistency in an unconstrained predictor is sufficient.


\subsection{Block Rotation JEPA (BRo-JEPA)}



The main contribution of this work is BRo-JEPA, a JEPA-style latent world model with an operation-conditioned block-rotation predictor. 
Compositional baselines rely on an unconstrained MLP predictor, which can fit seen operations without necessarily preserving the latent geometry required for strict zero-shot operation generalization. In contrast, BRo-JEPA builds the modulo-compositional structure directly into the predictor by representing arithmetic operations as rotations in latent space.

Our block-rotation predictor is inspired by Rotary Position Embedding (RoPE) \cite{su2021roformer}. Unlike standard uses of RoPE as a positional encoding, we use RoPE-inspired rotations as the latent dynamics operator itself. This constrains the predictor to apply operation-dependent circular transformations, matching the cyclic structure of modulo arithmetic. Since modulo addition defines a circular latent space, applying operation $k$ should correspond to moving $k$ steps around the latent cycle. BRo-JEPA implements this principle by replacing the standard MLP predictor with a structured rotation operator, thereby enforcing compositional geometry rather than leaving it to be learned implicitly.





Similar to \cite{su2021roformer}, we use a block rotation, where each block is a 2D planar rotation:

$$R(\theta)=\begin{bmatrix}
\cos\theta & -\sin\theta \\
\sin\theta & \cos\theta
\end{bmatrix}$$
For a $2n$-dimensional latent space, the block rotation is:
$$R_{\text{block}}=
\begin{bmatrix}
R(\theta_1) & 0 & \cdots & 0 \\
0 & R(\theta_2) & \cdots & 0 \\
\vdots & \vdots & \ddots & \vdots \\
0 & 0 & \cdots & R(\theta_n)
\end{bmatrix}$$

The predictor is a function of the context latent vector $z$, operation $k$, and rotation $R_{\text{block}}$, defined as: 
$$g(z, k, R_{\text{block}}(\theta)) = (R_{\text{block}}(\theta))^{k}\cdot z $$  
For each 2D planar rotation we have $R(\theta)^{k} = R(k\theta)$. Thus, we can get the predicted latent $z_p$ as 
$$z_p = R_{\text{block}}(k\cdot \theta)\cdot z$$

The rotation angles $\theta_i$ can be set as fixed parameters or can also be learnable parameters for our block rotation predictor. In each case, we tested two different strategies: Single-Frequency Rotation (SFR) and Multi-Frequency Rotation (MFR). In SFR each block (2D plane in the latent space) shares the same rotation angle $\theta$ while in MFR each block uses its individual $\theta$ as shown in Fig. \ref{fig:our_jepa_figure}.
   
For fixed rotation angles, the modular inductive bias is applied by setting $\theta = 2\pi/N$ in SFR and $\theta_i =  2\pi/N * (i \bmod(N/2))$ in MFR, where $N$ is the class count, $N=10$ for MNIST and $N=26$ for EMNIST (illustrated in Appx. Fig. \ref{fig:mfr_demo}).
The learnable-angle variants relax the requirement that the rotation period itself be specified. In these models, we retain the inductive assumption that the latent dynamics are rotational, but initialize $\theta_i$ randomly and optimize them from the primitive operations $\{-1,+1\}$. As shown in Fig.~\ref{fig:supple_lc}, the learned angles converge to frequencies compatible with the underlying modulo-10 period.

\subsection{Representation Collapse Mitigation}
\begin{figure}[t]
\centering
   \includegraphics[width=0.8\linewidth]{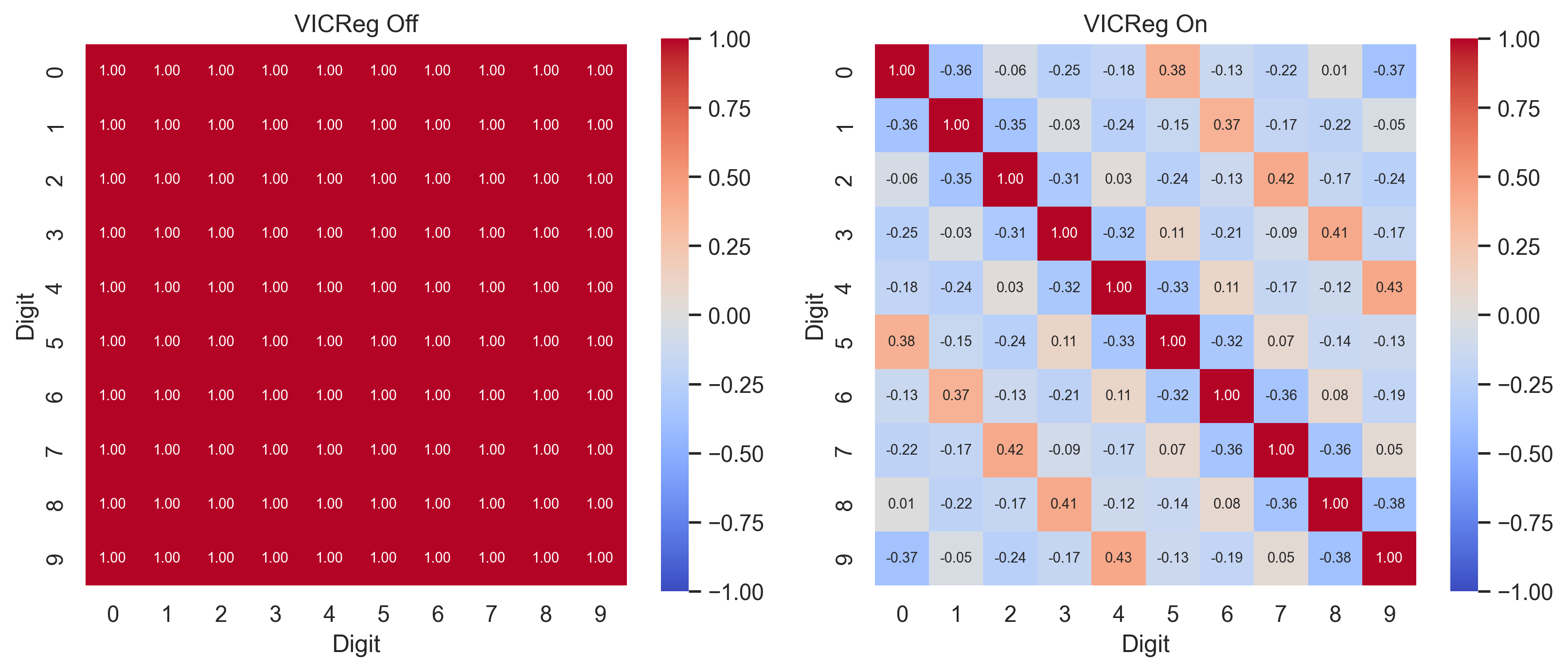}
   \caption{VICReg mitigates latent collapse: without VICReg, all class prototypes collapse to cosine similarity $1$; with VICReg, off-diagonal similarities move closer to $0$, indicating better-separated class prototypes.}   
\label{fig:collapse}
\end{figure}

JEPA models can exhibit a degenerate solution known as representation collapse \cite{assran2023self}, where different inputs map to low-variance or indistinguishable latents; we refer to this as latent collapse. This happens in JEPA training when optimization is solely based on the L2 loss between predicted and target latent representations.
To mitigate this, several strategies have been proposed. Following the JEPA framework, we use a non-gradient target encoder updated as an exponential moving average (EMA) of the context encoder. Additionally, we also use Variance-Invariance-Covariance Regularization (VICReg) \cite{bardes2021vicreg} that explicitly regularizes the latent space, enforces sufficient feature variance, and reduces redundancy. We adopted VICReg to stabilize training and prevent latent collapse, see Fig. \ref{fig:collapse} (details in Appx. \ref{sec:vicreg}).
\label{sec:latent_collapse}

 


\section{Training Details}
\label{sec:opt_and_training}

We select checkpoints using best held-out validation loss on held-out context images and seen operations. Zero-shot operation accuracy is reported on the held-out test operations. Models were trained in PyTorch \cite{paszke2019pytorch} for 25 epochs using AdamW \cite{loshchilov2019decoupled} with learning rate $0.0001$, unless otherwise stated.

We train supervised baselines with a 2-layer MLP with 256 hidden units and ResNet-18 backbones using cross-entropy loss. The MLP baseline takes a flattened MNIST image concatenated with an operation embedding, while the ResNet-18 baseline first encodes the image and then concatenates the resulting feature vector with the operation embedding before classification \cite{he2016deep}.


For a fair comparison with BRo-JEPA, which uses a block-rotation predictor, we also add block rotation to the supervised baselines.
In these models, the input image is first mapped to a feature representation by the MLP or ResNet-18 encoder. An operation-dependent block rotation is then applied to this representation before the final classifier head. The model is then trained using a cross-entropy loss.
For MNIST, we adapt torchvision's ResNet-18 by replacing the initial $7\times7$ stride-2 convolution with a $3\times3$ stride-1 convolution and removing the max-pooling layer. For both of the architectures, the size of latent dimension is 64.

\textbf{Dataset and Setup: }We formulate the task as latent prediction over MNIST, where each image represents a state and arithmetic operations define transitions. Given a context image $x_{\text{ctx}}$ and an operation $k \in \{-9, \dots, 9\}$, the model predicts the latent of the target image $x_{\text{tgt}}$. All images are normalized using standard MNIST statistics.
\label{sec:dataset_setup}




To generate transition pairs, we group images by label and sample targets using modulo arithmetic. Let $y_{\text{ctx}} \in \{0, \dots, 9\}$ be the label of $x_{\text{ctx}}$, and define
$y_{\text{tgt}} = (y_{\text{ctx}} + k) \bmod 10.$
We then uniformly sample $x_{\text{tgt}}$ from the subset with label $y_{\text{tgt}}$. Context and target images are sampled from the current split (train or test), and $k$ is sampled uniformly from the split-specific set of operations.

We define \textit{strict zero-shot operation generalization} as evaluation on operation values that are never seen 
during training. Unless otherwise stated, models are trained only on operations $\{-1,+1\}$ and evaluated on unseen operations $\{\pm2,\ldots,\pm9\}$. Models trained with compositional consistency loss use composed operations and are therefore reported as \textit{weak zero-shot} rather than strict.



\section{Measuring Performance}
We evaluate models primarily using classification accuracy on the MNIST test set, as the dataset is class-balanced. For supervised baselines, accuracy is computed directly from output logits. For JEPA-based models, accuracy is computed using the probe classifier attached to the predicted latent representation (Fig.~\ref{fig:our_jepa_figure}).

Table~\ref{tab:operation_embedding_ablation} reports train, seen-operation , zero-shot operation, rollout, and latent-space nearest-neighbor accuracies. Seen-operation accuracy uses the MNIST test split with the train operations. Zero-shot accuracy evaluates unseen operations, and nearest-neighbor accuracy checks whether predicted latents are class-discriminative without relying only on the classifier head.

For JEPA models, rollout accuracy measures compositional consistency by repeatedly applying primitive operations in latent space. For example, starting from digit $0$, applying $+1$ five times should produce a latent classified as digit $5$. We do not report rollout accuracy for supervised baselines because they output logits directly and do not define iterative latent dynamics.

Finally, we analyze latent structure using class prototypes, computed as the mean latent vector per digit class. Cosine-similarity matrices between these prototypes are used to visualize class separation (Figs.~\ref{fig:collapse} and \ref{fig:heatmap_comparison}).


\section{Experiments and Results}

We conducted several quantitative and qualitative experiments to evaluate model performance. A summary of our quantitative results is presented in Table \ref{tab:operation_embedding_ablation}.

\begin{table*}[t]
\caption{Ablation over model backbone, compositional consistency loss (c.c. loss), rotation variant (learnable or fixed), and training objective. The train operations used in this table are just the base operations [+1, -1]. The test operations are [$\pm$2..$\pm$9]. Rollout accuracy measures multi-step operation consistency. Values are reported as mean ± std over 6 seeds (0–5). Checkpoints are selected by best held-out validation loss, except JEPA + additive-embedding rows, reported at final epoch of a fixed run (25 epochs; ResNet18 w/ compositional-consistency loss: 200 epochs).}
\centering
\setlength{\tabcolsep}{5pt}
\resizebox{\textwidth}{!}{%
\begin{tabular}{lccccc}
\toprule
\textbf{Method} & \textbf{Train Acc.} & \textbf{Seen Op Acc.} & \textbf{Zero-Shot Op Acc.} & \textbf{Unseen Op Rollout Acc.} & \textbf{Unseen Op KNN Acc.} \\
\midrule

\multicolumn{6}{l}{\textit{\textbf{Supervised baselines + additive embedding}}} \\
MLP & 0.9888 $\pm$ 0.0073 & 0.9734 $\pm$ 0.0034 & 0.0521 $\pm$ 0.0072 & -- & 0.0353 $\pm$ 0.0045 \\
ResNet18 & 0.9809 $\pm$ 0.0119 & 0.9793 $\pm$ 0.0114 & 0.0127 $\pm$ 0.0081 & -- & 0.0012 $\pm$ 0.0006 \\

\midrule
\multicolumn{6}{l}{\textit{\textbf{JEPA + additive embedding}}} \\
MLP & 0.9960 $\pm$ 0.0003 & 0.9776 $\pm$ 0.0005 & 0.0265 $\pm$ 0.0068 & 0.9775 $\pm$ 0.0008 & 0.0263 $\pm$ 0.0079 \\
MLP w/ compositional consistency loss (weak zero shot) & 0.9932 $\pm$ 0.0010 & 0.9771 $\pm$ 0.0019 & 0.9403 $\pm$ 0.0143 & 0.9768 $\pm$ 0.0019 & 0.9398 $\pm$ 0.0160 \\
ResNet18 & 0.9983 $\pm$ 0.0004 & 0.9932 $\pm$ 0.0005 & 0.0348 $\pm$ 0.0232 & 0.9932 $\pm$ 0.0008 & 0.0327 $\pm$ 0.0247 \\
ResNet18 w/ compositional consistency loss (weak zero shot) & 0.9995 $\pm$ 0.0002 & 0.9946 $\pm$ 0.0007 & 0.9285 $\pm$ 0.0246 & 0.9948 $\pm$ 0.0007 & 0.9400 $\pm$ 0.0225 \\

\midrule
\midrule
\multicolumn{6}{l}{\textit{\textbf{Supervised baselines + block rotation on features of the encoder}}} \\
MLP w/ single-frequency rotation (SFR, fixed) & 0.9683 $\pm$ 0.0015 & 0.9622 $\pm$ 0.0019 & 0.1439 $\pm$ 0.0134 & -- & 0.0697 $\pm$ 0.0320 \\
MLP w/ single-frequency rotation (SFR, learnable) & 0.9893 $\pm$ 0.0123 & 0.9745 $\pm$ 0.0074 & 0.1418 $\pm$ 0.0018 & -- & 0.1338 $\pm$ 0.0036 \\
MLP w/ multi-frequency rotation (MFR, fixed) & 0.9952 $\pm$ 0.0068 & 0.9785 $\pm$ 0.0031 & 0.5454 $\pm$ 0.1047 & -- & 0.5207 $\pm$ 0.1114 \\
MLP w/ multi-frequency rotation (MFR, learnable) & 0.9922 $\pm$ 0.0064 & 0.9775 $\pm$ 0.0029 & 0.3150 $\pm$ 0.0531 & -- & 0.2651 $\pm$ 0.0457 \\
ResNet18 w/ single-frequency rotation (SFR, fixed) & 0.9851 $\pm$ 0.0025 & 0.9850 $\pm$ 0.0030 & 0.1709 $\pm$ 0.0511 & -- & 0.1613 $\pm$ 0.0282 \\
ResNet18 w/ single-frequency rotation (SFR, learnable) & 0.9814 $\pm$ 0.0068 & 0.9801 $\pm$ 0.0079 & 0.1275 $\pm$ 0.0167 & -- & 0.1208 $\pm$ 0.0105 \\
ResNet18 w/ multi-frequency rotation (MFR, fixed) & 0.9981 $\pm$ 0.0011 & 0.9935 $\pm$ 0.0004 & 0.5362 $\pm$ 0.0476 & -- & 0.4975 $\pm$ 0.0353 \\
ResNet18 w/ multi-frequency rotation (MFR, learnable) & 0.9960 $\pm$ 0.0040 & 0.9922 $\pm$ 0.0031 & 0.5027 $\pm$ 0.0564 & -- & 0.4805 $\pm$ 0.0659 \\

\midrule
\multicolumn{6}{l}{\textit{\textbf{BRo-JEPA}}} \\
MLP w/ single-frequency rotation (SFR, fixed) & 0.9690 $\pm$ 0.0069 & 0.9426 $\pm$ 0.0026 & 0.9418 $\pm$ 0.0026 & 0.9419 $\pm$ 0.0028 & 0.9421 $\pm$ 0.0030 \\
MLP w/ single-frequency rotation (SFR, learnable) & 0.9707 $\pm$ 0.0050 & 0.9430 $\pm$ 0.0023 & 0.9424 $\pm$ 0.0031 & 0.9428 $\pm$ 0.0031 & 0.9425 $\pm$ 0.0029 \\
MLP w/ multi-frequency rotation (MFR, fixed) & \textbf{0.9936 $\pm$ 0.0011} & 0.9724 $\pm$ 0.0014 & \textbf{0.9724 $\pm$ 0.0009} & 0.9722 $\pm$ 0.0011 & \textbf{0.9727 $\pm$ 0.0011} \\
MLP w/ multi-frequency rotation (MFR, learnable) & 0.9924 $\pm$ 0.0015 & \textbf{0.9736 $\pm$ 0.0014} & \textbf{0.9724 $\pm$ 0.0012} & \textbf{0.9723 $\pm$ 0.0016} & 0.9713 $\pm$ 0.0018 \\
ResNet18 w/ single-frequency rotation (SFR, fixed) & 0.9914 $\pm$ 0.0023 & 0.9891 $\pm$ 0.0013 & 0.9891 $\pm$ 0.0014 & 0.9890 $\pm$ 0.0014 & 0.9891 $\pm$ 0.0016 \\
ResNet18 w/ single-frequency rotation (SFR, learnable) & 0.9907 $\pm$ 0.0023 & 0.9895 $\pm$ 0.0012 & 0.9894 $\pm$ 0.0014 & 0.9894 $\pm$ 0.0013 & 0.9893 $\pm$ 0.0013 \\
ResNet18 w/ multi-frequency rotation (MFR, fixed) & 0.9989 $\pm$ 0.0006 & 0.9942 $\pm$ 0.0006 & 0.9942 $\pm$ 0.0005 & 0.9941 $\pm$ 0.0007 & 0.9942 $\pm$ 0.0007 \\
ResNet18 w/ multi-frequency rotation (MFR, learnable) & \textbf{0.9992 $\pm$ 0.0002} & \textbf{0.9944 $\pm$ 0.0006} & \textbf{0.9944 $\pm$ 0.0004} & \textbf{0.9944 $\pm$ 0.0006} & \textbf{0.9944 $\pm$ 0.0006} \\

\bottomrule
\end{tabular}%
}
\label{tab:operation_embedding_ablation}
\end{table*}

\subsection{Baseline Results}

\textbf{Classification baselines capture the sign of the operation but fail to encode the magnitude: } The baseline models are described in Section \ref{sec:opt_and_training}.
From Table \ref{tab:operation_embedding_ablation}, we see that the models perform exceptionally on seen operations reaching high accuracies of 97.34\% and 99.93\% for MLP and ResNet-18 variants. However, the zero-shot and unseen operation accuracy maxed to only 5.21\% and 1.27\% respectively. This shows that the model fails to generalize even when using additive embedding, which had the extrapolation bias in the embeddings.



\begin{figure}[t!]
  \centering
  \includegraphics[width=0.5\linewidth]{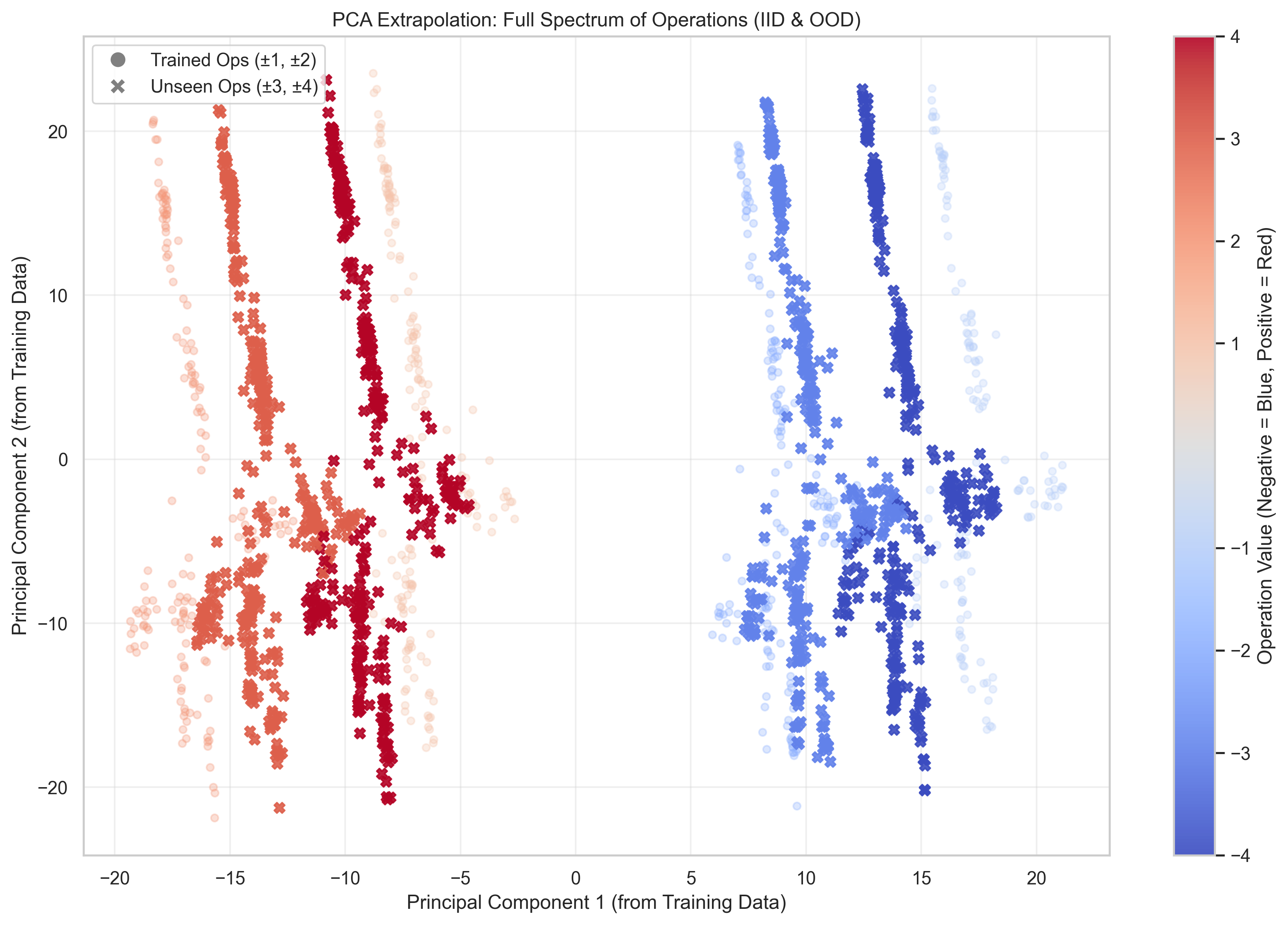}
  \caption{PCA visualization of ResNet baseline latent representation reveals the model captures the sign of operations (red vs. blue) but fails to encode the magnitude (dot vs. cross). Unseen operations ($\pm 3, \pm 4$) arbitrarily crammed in the in-distribution representations($\pm 1, \pm 2$).}
  \label{fig:baseline_extrapolation}
\end{figure}

The Principal Component Analysis (PCA) visualization of the ResNet baseline's latent space (Fig. \ref{fig:baseline_extrapolation}) provides insight into why it fails to generalize to unseen operations. The model partitions its latent representation into operation-specific tracks on the seen operations $\pm 1, \pm2$, forming 4 distinct linear structures (light red and blue). However, when presented with out-of-distribution operations such as $\pm 3, \pm4$, the model correctly predicts the sign of operations but fails to encode the magnitude.  As a result, unseen operations are arbitrarily crammed into the in-distribution representation space, leading to extremely poor accuracy. This behavior suggests that baseline lacks the continuous and compositional structural representation required to interpolate or extrapolate the correct target class, resulting in predictive collapse.

\begin{figure}[t]
\includegraphics[width=\linewidth]{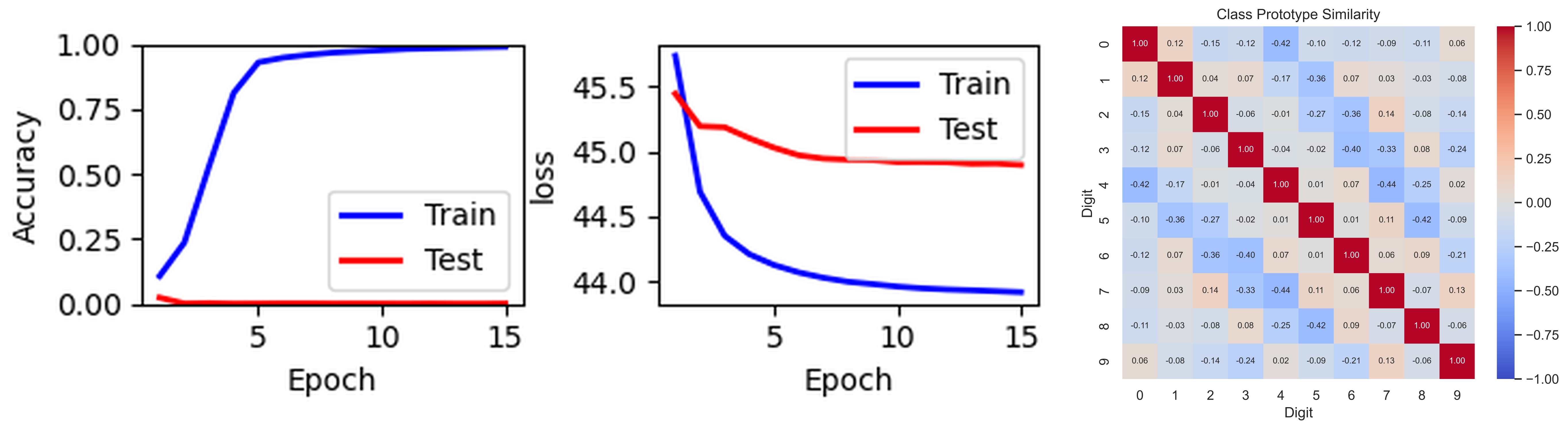}
   \caption{Learning curves show the accuracy and loss during training and test for JEPA model. Class prototype similarity matrix shows the trained target latent cosine similarity for MLP predictor (train op $\pm 1, \pm2, \pm4$ and test ops $\pm 3$) }
\label{fig:lc_jepa}
\end{figure}

\textbf{JEPA learns symbolic arithmetic operations; fails to extrapolate: }Training metrics in Table \ref{tab:operation_embedding_ablation} indicate that the model learned the seen operations effectively: as the training epoch increases, the training loss decreases and the accuracy increases to 98.88\% (Fig. \ref{fig:lc_jepa}).


Despite strong performance on seen operations, the model fails to generalize to unseen operations, with high test loss and near-zero accuracy (left and middle, Fig. \ref{fig:lc_jepa}). The prototype similarity matrix  (right, Fig. \ref{fig:lc_jepa}) shows a weak circular structure where similarity decreases with digit distance before increasing. This suggests that, unlike the baseline model, typical JEPA can capture some compositional and periodic structure in the digits, but the signal is too weak to emerge.

\textbf{Additive action embeddings and compositional consistency loss do not guarantee strict zero-shot generalization} Additive action embeddings encode compositional structure in the operation representation, but Table~\ref{tab:operation_embedding_ablation} shows that this alone is not sufficient for strict zero-shot generalization. JEPA models with additive embeddings perform well on seen operations, reaching $97.76\%$ seen-operation accuracy with an MLP and $99.32\%$ with ResNet-18. They also achieve high rollout accuracy, $97.75\%$ and $99.32\%$, showing that the learned primitive transitions are internally consistent.

However, direct one-step prediction on unseen composed operations remains poor: zero-shot accuracy is only $2.65\%$ for MLP and $3.48\%$ for ResNet-18, with similarly low KNN accuracies of $2.63\%$ and $3.27\%$. This suggests that although the embedding encodes the correct arithmetic structure, the unconstrained nonlinear predictor does not reliably preserve it for unseen operations.

Adding compositional consistency loss improves weak zero-shot accuracy to $94.03\%$ for MLP and $92.85\%$ for ResNet-18. However, this is not strict zero-shot because composed operations are indirectly used during training through the auxiliary loss. These results motivate BRo-JEPA, where the modular compositional structure is built directly into the predictor.

\textbf{Primitive latent dynamics support rollout but not direct composition: } Although additive-embedding JEPA fails at direct one-step zero-shot prediction, its learned primitive operations remain highly stable under rollout. When the model recursively applies the learned $+1$ or $-1$ transition to produce higher-order operations such as $+n$ or $-n$, rollout accuracy remains high: $97.75\%$ for the MLP backbone and $99.32\%$ for ResNet-18. Similar rollout performance is observed with compositional consistency loss, reaching $97.68\%$ for MLP and $99.48\%$ for ResNet-18.

This shows that the model learns reliable local latent transitions for primitive operations. However, this local consistency does not automatically translate into direct one-step inference of unseen composed operations. In other words, repeatedly applying $+1$ can approximate $+3$, but giving the model $+3$ directly still fails unless the predictor is explicitly constrained or trained to preserve the compositional structure.

\subsection{BRo-JEPA Results }

\textbf{Fixed Rotation Angles: }
In the fixed-angle setting, the modulo structure is explicitly encoded through the rotation angles; therefore, compositionality of the transition operator is ensured by construction. The remaining learning problem is to organize visual representations so that the imposed rotations correspond to the correct semantic digit transitions. Under this setting, BRo-JEPA achieves zero-shot unseen-operation accuracies of of 94.18\% and 98.91\% (SFR); 97.24\% and 99.42 \% (MFR) for MLP and ResNet encoders respectively (Table \ref{tab:operation_embedding_ablation}).


\begin{figure}[t]
  \centering

  \begin{minipage}[t]{0.48\linewidth}
    \centering
    \includegraphics[width=\linewidth]{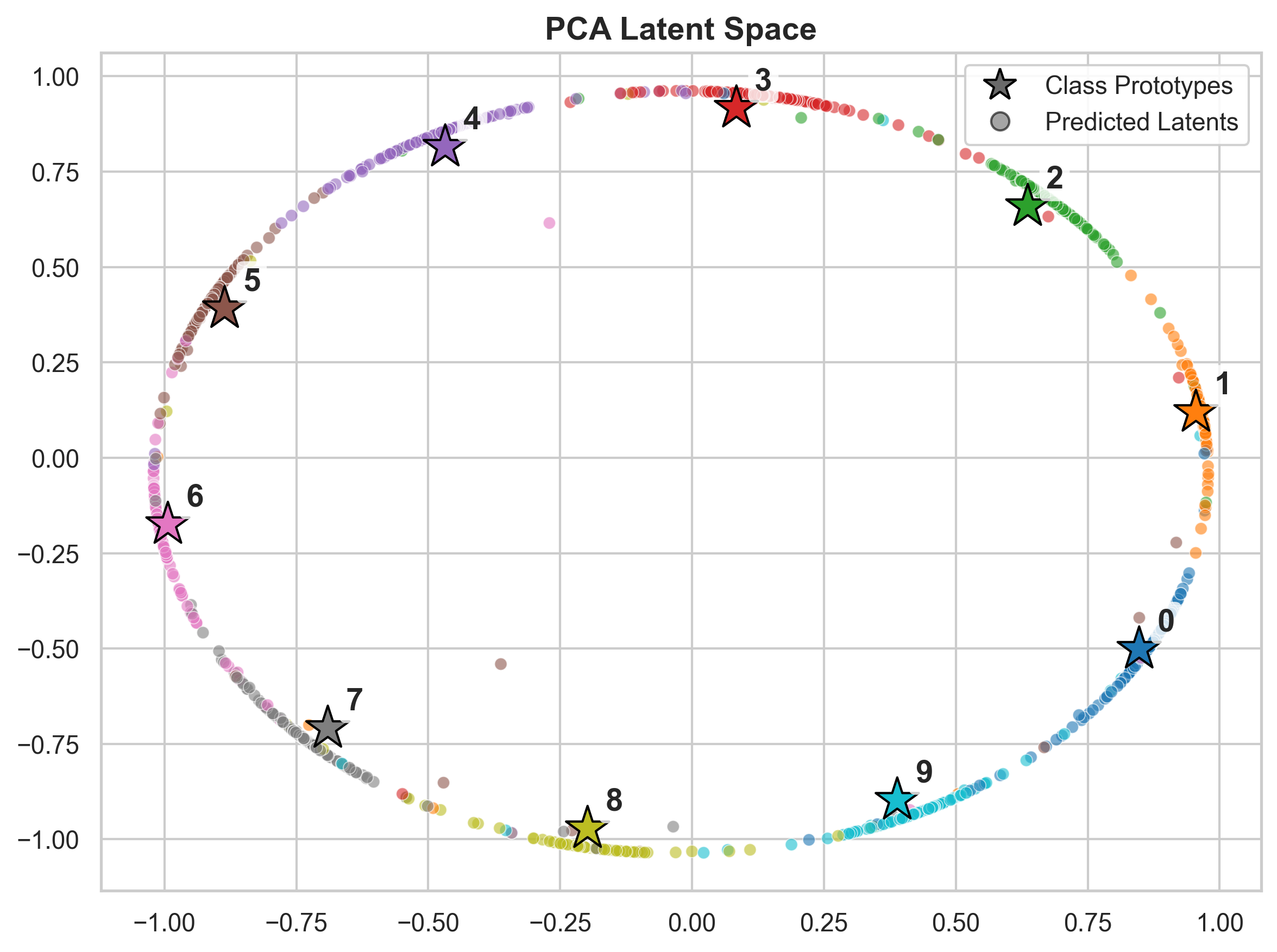}
    \captionof{figure}{Latent space PCA of SFR block-rotation predictor. By explicitly enforcing rotational dynamics, BRo-JEPA successfully disentangles the latent space into a circular manifold representing the modulo-10 arithmetic structure.}
    \label{fig:blockrot_pca}
  \end{minipage}
  \hfill
  \begin{minipage}[t]{0.48\linewidth}
    \centering
    \includegraphics[width=\linewidth]{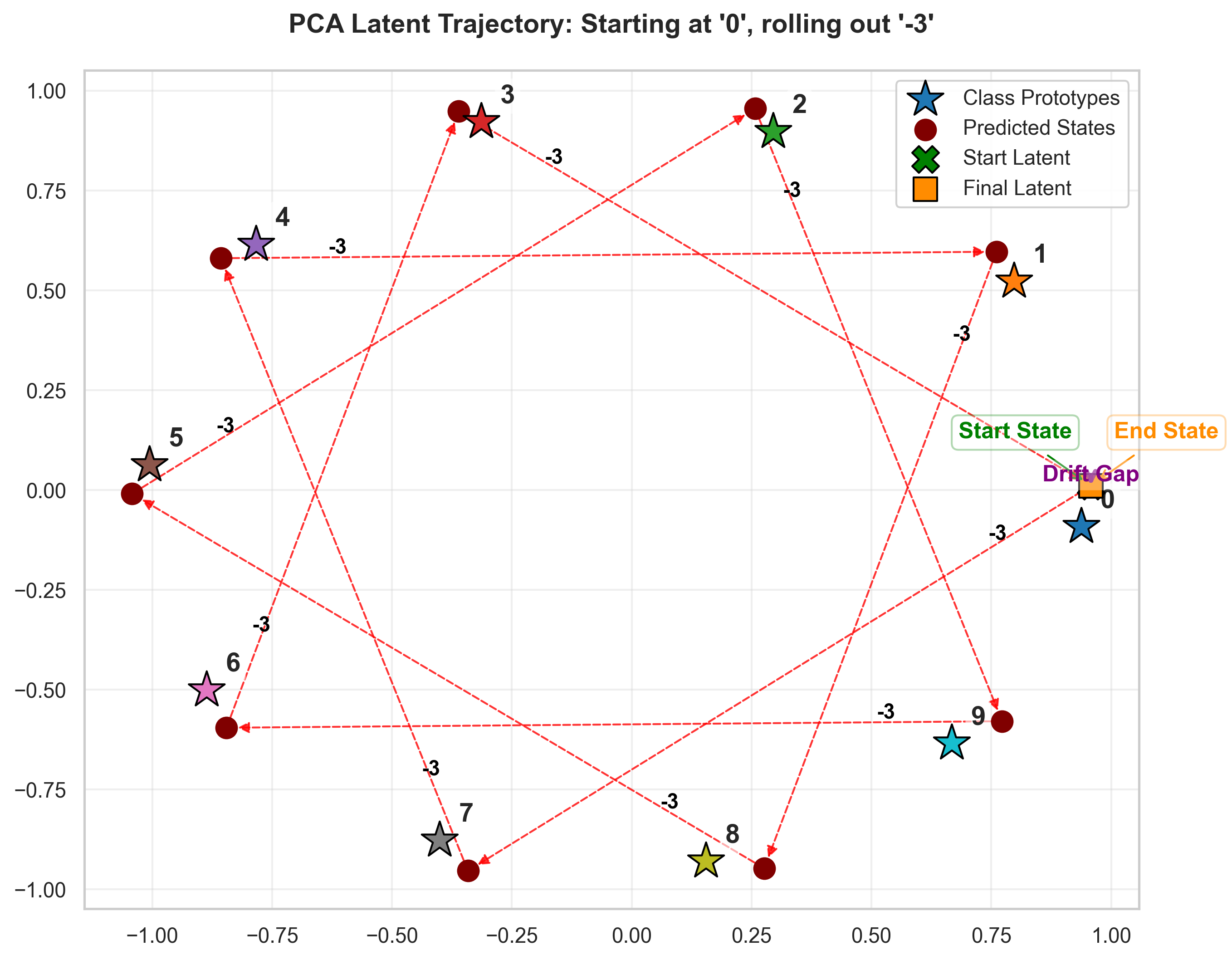}
    \captionof{figure}{PCA visualization of multi-step latent trajectories for the SFR block-rotation predictor under the zero-shot operation $-3$. Starting from digit $0$, the model composes unseen operations with no drift across rollout steps.}
    \label{fig:traj_minus3}
  \end{minipage}

\end{figure}

\textbf{Geometric Structure of Latent Space in SFR: } The PCA of image representations shows that block rotation successfully coerces the latent space into a continuous elliptical manifold in SFR (Fig. \ref{fig:blockrot_pca}). The class prototypes arrange themselves sequentially on the circle. This geometric configuration demonstrates that the model internalized the modular arithmetic topology that corresponds latent rotation to arithmetic operations. 

\begin{figure}[t]
\centering
   \includegraphics[width=0.8\linewidth]{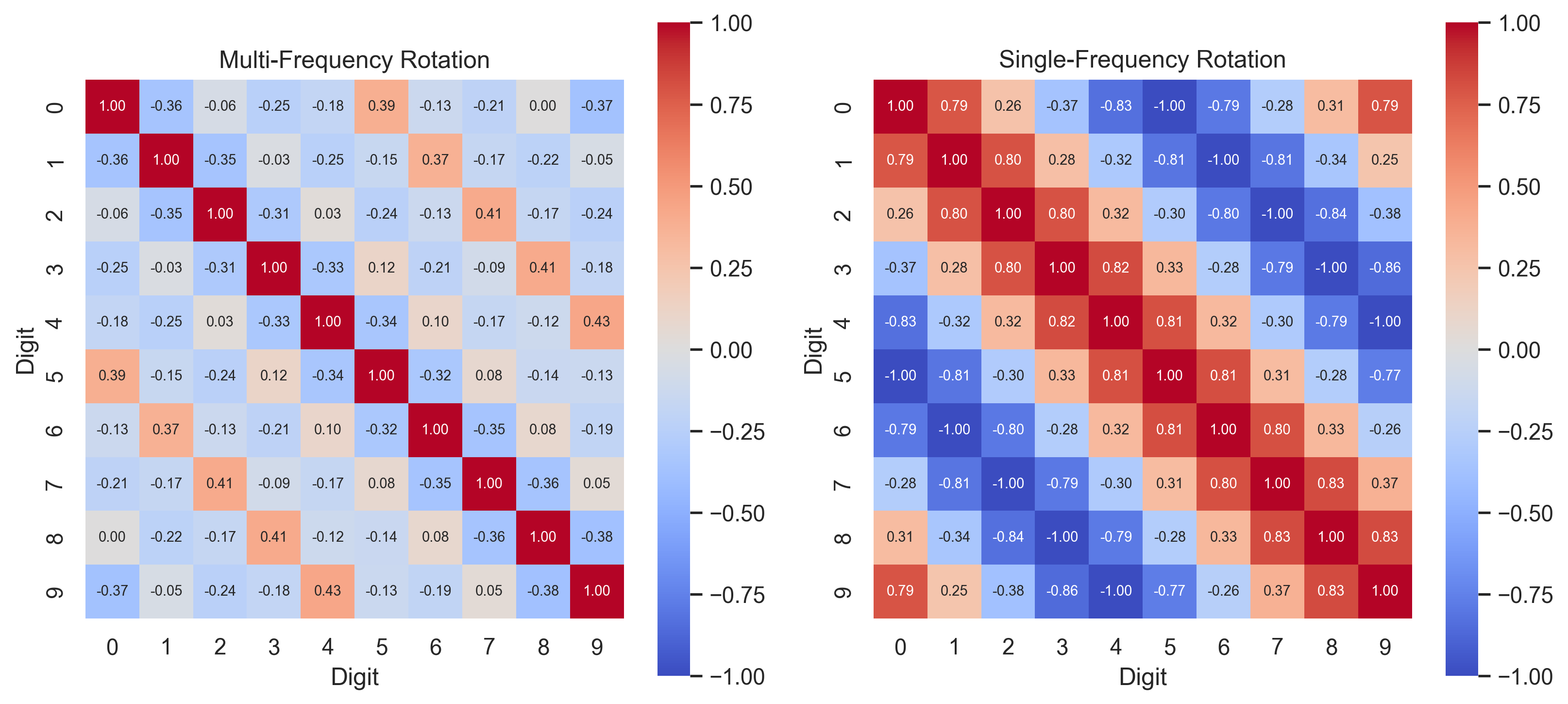}
   \caption{Class prototype cosine-similarity matrices for MFR and SFR block-rotation JEPA.}
\label{fig:heatmap_comparison}

\end{figure}

In the SFR latent space, the fixed rotation angle $\theta = 2\pi/10$ induces a circular modulo-10 structure. Therefore, digit pairs five steps apart, such as $0/5$, $1/6$, $2/7$, $3/8$, and $4/9$, become approximately antipodal because applying $+5$ corresponds to a rotation by $\pi$. This explains the near $-1$ cosine similarity for these pairs in Fig.~\ref{fig:heatmap_comparison}. Pairs four or six steps apart, such as $0/4$ or $1/5$, also show strong negative similarity because their angular separation is close to, but not exactly, $\pi$. In PCA, these negatively correlated prototypes appear on opposite sides of the circular manifold, reflecting the learned modulo-10 geometry. The corresponding MFR PCA visualization is shown in Appx. Fig.~\ref{fig:blockrot_pca_MFR}.

In Fig.~\ref{fig:heatmap_comparison}, MFR has off-diagonal similarities closer to zero than SFR, showing that its class prototypes are more decorrelated and separated. While SFR uses one rotation frequency and creates strong periodic correlations, MFR spreads the modulo-10 structure across multiple frequencies. This prevents any single circular pattern from dominating the similarity matrix and gives a balanced latent space. MFR's higher accuracy (Table \ref{tab:operation_embedding_ablation}) confirms that MFR better captures underlying structure and improve robustness to dataset variability.



PCA (Fig. \ref{fig:blockrot_pca}) shows the circular latent manifold; multi-step rollouts test its stability by repeatedly applying an operation from digit $0$. The -3 trajectory (Fig. \ref{fig:traj_minus3}) highlights the model's extrapolation and compositional characteristics.  Although the model was trained only on $\pm 1$, it follows the correct zero-shot sequence $0 \rightarrow 7 \rightarrow 4 \dots$ when given the $-3$ operation. Low drift suggests that the predictor learned the modular arithmetic rule rather than memorizing local transitions. More trajectories are in Appx.~Fig.~\ref{fig:trajectories_sup}.



\textbf{Learnable Rotation Angles: }
We next test whether the rotational frequencies can be recovered from the training transitions rather than specified using the known modulus. In these experiments, the predictor retains the block-rotation parameterization, but the rotation angles are randomly initialized and optimized from training data containing only the primitive operations $\{-1,+1\}$. In particular, the value $2\pi/10$ is not supplied to or used to constrain the learnable angles.

\begin{figure}[t]
\centering
\includegraphics[width=0.7\linewidth]{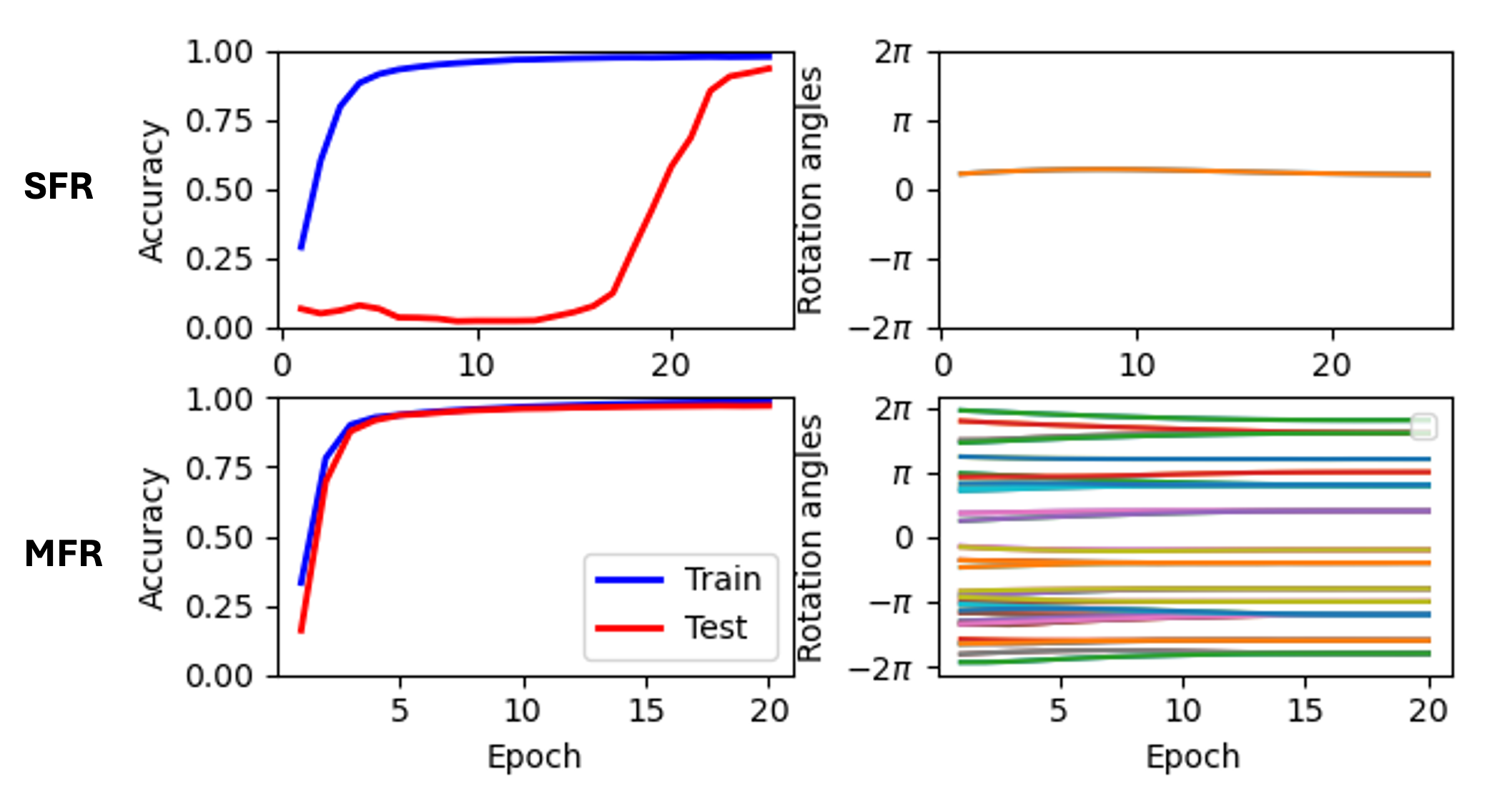}
   \caption{Learning curves show the accuracy (left) and rotation angles (right) during training for SFR and MFR}
\label{fig:lc_block_rotation}
\end{figure}

In both SFR and MFR the predictor successfully learns the rotation angles when $\theta$s are learnable parameters, converging rapidly during training to accuracies of $97.07\%$ and $99.24\% $ accordingly (Table \ref{tab:operation_embedding_ablation}, MLP section). However, generalization to unseen operations differs remarkably. In MFR, zero shot accuracy quickly matches the training performance with similar trajectory, achieving $97.36\%$. In contrast, SFR exhibits a significant delay (see Fig \ref{fig:lc_block_rotation}, left): the testing accuracy remains near 0 for many epochs then gradually reaches $94.24\%$, way below the training performance. This indicates while SFR learns representations to fit the training data but fails to generalize early on, MFR facilitates learning of effectively-generalized representations at the outset.

Starting from random initialization in the range of $[-2\pi, 2\pi]$, the learned rotation angles converge to discrete values corresponding to $\pi/5$ and its harmonics as shown in Fig. \ref{fig:lc_block_rotation} (right). In SFR, the shared rotation angle $\theta$ converges to $\pi/5$ (left), aligned with that in our fixed rotation angles. In MFR, the 32 learned rotation angles converge to a set of distinct harmonic components of $\pi/5$, with 13 unique ones observed here.

The specific harmonic component to which each rotation angle converges depends on the initial conditions. The current JEPA model objective exhibits multiple local minima corresponding to these harmonic solutions. As a result, the learned rotation angles tend to be trapped in one of these minima. This behavior affects MFR little due to the presence of multiple independent rotation angles, but it significantly degrades the performance in SFR when the learned angle converges to a higher order harmonic of $\pi/5$ instead of the fundamental frequency itself (see Appx. \ref{sec: sfr_degrades}).  


BRo-JEPA captures the modulo-10 structure in the data by leveraging a compositional inductive bias in the latent space under both SFR and MFR. By allowing independent rotation angles, MFR learns a Fourier-like decomposition of transformations, where different latent pairs encode different frequency components of the modulo-10 dynamics. This suggests that the learned latent dynamics distribute modular structure across multiple frequency components rather than relying on a single global rotation, which helps explain MFR's faster convergence and stronger generalization to unseen operations.




\textbf{Zero-Shot Performance on Unseen Operations:}
Although the compositional consistency loss achieves high zero-shot accuracy of $92.85\%$ (Table \ref{tab:operation_embedding_ablation}, ResNet), it should be considered weakly zero-shot because the predictor is indirectly exposed to composed operations through the auxiliary training objective. In contrast, the block-rotation predictor is strictly zero-shot with respect to operations: the model is trained only on the base operations $(+1,-1)$, and all unseen operations are inferred through the rotational structure imposed in latent space. Its $99.44\%$ zero-shot accuracy demonstrates that BRo-JEPA aligns its latent representation with the modular arithmetic structure, enabling generalization to unseen arithmetic operations.

\textbf{Block rotation alone does not explain BRo-JEPA's zero-shot performance:} To separate the effect of the structured rotation operator from the JEPA learning objective, we compare BRo-JEPA against supervised baselines using the same block-rotation operator. Although these supervised models achieve high seen-operation accuracy, they generalize substantially worse to unseen operations. For ResNet-18 with MFR, the supervised model achieves $53.62\%$ zero-shot accuracy with fixed angles and $50.27\%$ with learnable angles, compared with $99.42\%$ and $99.44\%$, respectively, for BRo-JEPA. Thus, constraining features with the correct transformation family is not by itself sufficient for semantic zero-shot generalization; the JEPA objective encourages the visual latent representation to become compatible with the structured dynamics.

\subsection{BRo-JEPA on EMNIST Letters}

We extend our experiments to EMNIST Letters, a 26-class, class-balanced dataset of handwritten letters split into 124{,}800 training and 20{,}800 test images~\cite{cohen2017emnist}. Upper and lower case instances of a letter share the same label, so within-class variability is substantially higher than in MNIST.

As with MNIST, arithmetic operations are defined cyclically over the label space. For example, the operation $+1$ maps ``a'' $\rightarrow$ ``b'', with modular wraparound such that ``z'' $\rightarrow$ ``a''.

Table \ref{tab:operation_embedding_ablation_mean_sd} shows a similar pattern seen on MNIST for EMNIST Letters, though with lower overall accuracy given the higher complexity of the dataset. BRo-JEPA achieves higher accuracy on strict zero-shot evaluation.

SFR trails much further behind MFR on EMNIST compared to MNIST. SFR reaches only 51.27\% zero-shot accuracy, compared to 94.38\% for MFR. This suggests a single shared rotation frequency has less capacity to accommodate 26 classes, whereas MFR is able to distribute the modulo-26 structure across multiple independent frequencies.  

MFR's use of multiple frequencies provides additional degrees of freedom to maintain class separability even as within-class visual variability (upper and lower case forms) increases. These results suggest that MFR's advantage on MNIST is not specific to the 10-class setting, but becomes more pronounced as the state space grows and within-class variability increases.


\begin{table*}[t]
\caption{EMNIST Letters ablations using the same protocol as Table \ref{tab:operation_embedding_ablation}.}
\centering
\resizebox{\textwidth}{!}{%
\begin{tabular}{lccccc}
\toprule
\textbf{Method} & \textbf{Train Acc.} & \textbf{Seen Op Acc.} & \textbf{Zero-Shot Op Acc.} & \textbf{Unseen Op Rollout Acc.} & \textbf{Unseen Op KNN Acc.} \\
\midrule

\multicolumn{6}{l}{\textit{\textbf{Supervised baselines + additive embedding}}} \\
MLP & 0.8436 $\pm$ 0.0145 & 0.8272 $\pm$ 0.0110 & 0.0374 $\pm$ 0.0038 & -- & 0.0295 $\pm$ 0.0034 \\
ResNet18 & 0.9306 $\pm$ 0.0185 & 0.9230 $\pm$ 0.0125 & 0.0133 $\pm$ 0.0010 & -- & 0.0048 $\pm$ 0.0005 \\

\midrule
\multicolumn{6}{l}{\textit{\textbf{JEPA + additive embedding}}} \\
MLP & 0.9747 $\pm$ 0.0003 & 0.9049 $\pm$ 0.0010 & 0.0209 $\pm$ 0.0039 & 0.9047 $\pm$ 0.0014 & 0.0202 $\pm$ 0.0024 \\
MLP w/ compositional consistency loss (weak zero shot) & 0.9625 $\pm$ 0.0023 & 0.8934 $\pm$ 0.0038 & 0.8347 $\pm$ 0.0289 & 0.8919 $\pm$ 0.0032 & 0.8449 $\pm$ 0.0207 \\
ResNet18 & 0.9959 $\pm$ 0.0005 & 0.9433 $\pm$ 0.0015 & 0.0296 $\pm$ 0.0031 & 0.9432 $\pm$ 0.0014 & 0.0274 $\pm$ 0.0022 \\
ResNet18 w/ compositional consistency loss (weak zero shot) & 0.9962 $\pm$ 0.0003 & 0.9433 $\pm$ 0.0011 & 0.7309 $\pm$ 0.0336 & 0.9436 $\pm$ 0.0013 & 0.7546 $\pm$ 0.0373 \\

\midrule
\midrule
\multicolumn{6}{l}{\textit{\textbf{Supervised baselines + block rotation on features of encoder}}} \\
MLP w/ single-frequency rotation (SFR, fixed) & 0.8264 $\pm$ 0.0409 & 0.8154 $\pm$ 0.0387 & 0.0894 $\pm$ 0.0054 & -- & 0.0821 $\pm$ 0.0076 \\
MLP w/ multi-frequency rotation (MFR, fixed) & 0.9049 $\pm$ 0.0178 & 0.8752 $\pm$ 0.0106 & 0.1318 $\pm$ 0.0100 & -- & 0.1087 $\pm$ 0.0122 \\
ResNet18 w/ single-frequency rotation (SFR, fixed) & 0.9552 $\pm$ 0.0302 & 0.9337 $\pm$ 0.0129 & 0.1085 $\pm$ 0.0093 & -- & 0.1063 $\pm$ 0.0100 \\
ResNet18 w/ multi-frequency rotation (MFR, fixed) & 0.9872 $\pm$ 0.0023 & \textbf{0.9450 $\pm$ 0.0017} & 0.2513 $\pm$ 0.0159 & -- & 0.2183 $\pm$ 0.0137 \\

\midrule
\multicolumn{6}{l}{\textit{\textbf{BRo-JEPA}}} \\
MLP w/ single-frequency rotation (SFR, learnable) & 0.4793 $\pm$ 0.0096 & 0.4007 $\pm$ 0.0071 & 0.2089 $\pm$ 0.0129 & 0.2089 $\pm$ 0.0124 & 0.2041 $\pm$ 0.0152 \\
MLP w/ multi-frequency rotation (MFR, fixed) & \textbf{0.9245 $\pm$ 0.0053} & \textbf{0.8469 $\pm$ 0.0034} & \textbf{0.6512 $\pm$ 0.0317} & \textbf{0.6509 $\pm$ 0.0344} & \textbf{0.5910 $\pm$ 0.0137} \\
MLP w/ multi-frequency rotation (MFR, learnable) & 0.8893 $\pm$ 0.0081 & 0.8323 $\pm$ 0.0040 & 0.5802 $\pm$ 0.0322 & 0.5794 $\pm$ 0.0358 & 0.5377 $\pm$ 0.0338 \\
ResNet18 w/ single-frequency rotation (SFR, learnable) & 0.7142 $\pm$ 0.1236 & 0.6883 $\pm$ 0.1118 & 0.5127 $\pm$ 0.0312 & 0.5123 $\pm$ 0.0312 & 0.4924 $\pm$ 0.0207 \\
ResNet18 w/ multi-frequency rotation (MFR, fixed) & 0.9984 $\pm$ 0.0001 & \textbf{0.9439 $\pm$ 0.0013} & \textbf{0.9438 $\pm$ 0.0012} & \textbf{0.9437 $\pm$ 0.0013} & \textbf{0.9439 $\pm$ 0.0013} \\
ResNet18 w/ multi-frequency rotation (MFR, learnable) & \textbf{0.9990 $\pm$ 0.0003} & 0.9437 $\pm$ 0.0015 & 0.9435 $\pm$ 0.0014 & 0.9432 $\pm$ 0.0014 & 0.9438 $\pm$ 0.0012 \\

\bottomrule
\end{tabular}%
}
\label{tab:operation_embedding_ablation_mean_sd}
\label{tab:emnist_table}
\end{table*}

\section{Limitations}
Our experiments are conducted on MNIST and EMNIST modular arithmetic, a controlled setting with a known cyclic group structure. Rather than suggesting that BRo-JEPA directly generalizes to arbitrary symbolic reasoning tasks, our results position it as a model for tasks where the underlying modular structure is known and can be built into the latent dynamics. In such settings, the block-rotation predictor provides a principled way to encode cyclic transformations and support zero-shot generalization over unseen operations. Future work should investigate this idea in broader modular visual domains, such as 2D image rotations, 3D object pose transformations, and other tasks where the relevant transformation structure is modular.


\section{Conclusion}

JEPA with additive operation embeddings achieves strong seen-operation and rollout performance but fails on unseen operations. BRo-JEPA addresses this by applying rotational transformations onto the latent representations, where the optimal rotation frequencies emerge from training on primitive operations without the need of an explicit modulo-$N$ prior. 
The performance gap between BRo-JEPA and supervised models using the same block-rotation operator indicates that the operator alone is not sufficient for zero-shot generalization. 
Together, these results suggest that combining structured latent transformations with predictive representation learning is a promising approach for tasks with a cyclic or modular structure.

\section*{Acknowledgments}
We thank Prof. Zsolt Kira for his guidance and feedback during the early development of this work. We also thank Revant Teotia for his feedback in the early stages of the project. 





\bibliographystyle{splncs04}
\bibliography{egbib_arxiv}

\clearpage

\appendix
\section{Supplementary Materials}

\subsection{Operation Embeddings}
\label{sec:op_embeddings}
We explored several ways of encoding the arithmetic operation as an input feature. The goal was to represent the operator in a form that the model can use for either memorization of seen operations or generalization to unseen ones.

\begin{itemize}
\item \textbf{Lookup embedding}: Each discrete operation is assigned a learned vector from an embedding table. This is the simplest option and works well when the set of operations is fixed and known in advance, but it does not naturally generalize to unseen values.

\item \textbf{Linear embedding}: The raw scalar operation value is projected through a small neural network. This gives a continuous representation that can handle arbitrary operation values and is more flexible than a lookup table.

\item \textbf{Fourier embedding}: The operation value is mapped to sinusoidal features such as 
$\sin(x),\ \cos(x),\ \sin(2x),\ \cos(2x)$
before projection. This encodes periodic structure and tends to support smoother generalization across nearby operations.

\item \textbf{Hybrid embedding}: The sign of the operation and its magnitude are encoded separately. The sign is handled with a categorical embedding, while the magnitude is represented with Fourier features. This aims to capture both operator direction and numeric scale.
\end{itemize}

\subsection{Training process for composition consistency loss}
\label{sec:composition_consistency_loss}
$z_{seq} (n*\text{primitive operation})$ is computed by recursively applying the primitive operation to the model n times, while $z_{dirct}(n*\text{primitive operation})$ is computed by directly applying the high order operation $n*\text{primitive operation}$ in one step.  

In practice, we find the smallest positive training op and smallest negative training op: +1 and -1 in the compositionality setup. Then for each requested composition op, for example 4 or -4, $z_{seq}$ is computed by applying the primitive operation 1 or -1 recursively 4 times. $z_{direct}$ is computed by applying 4 or -4 to the JEPA model directly. 
The composition consistency loss is calculated as $|z_{seq} -z_{direct}|^2$, which is added to the regular JEPA loss with a weight hyperparameter. 

\begin{figure}
    \centering
    \includegraphics[width=0.5\linewidth]{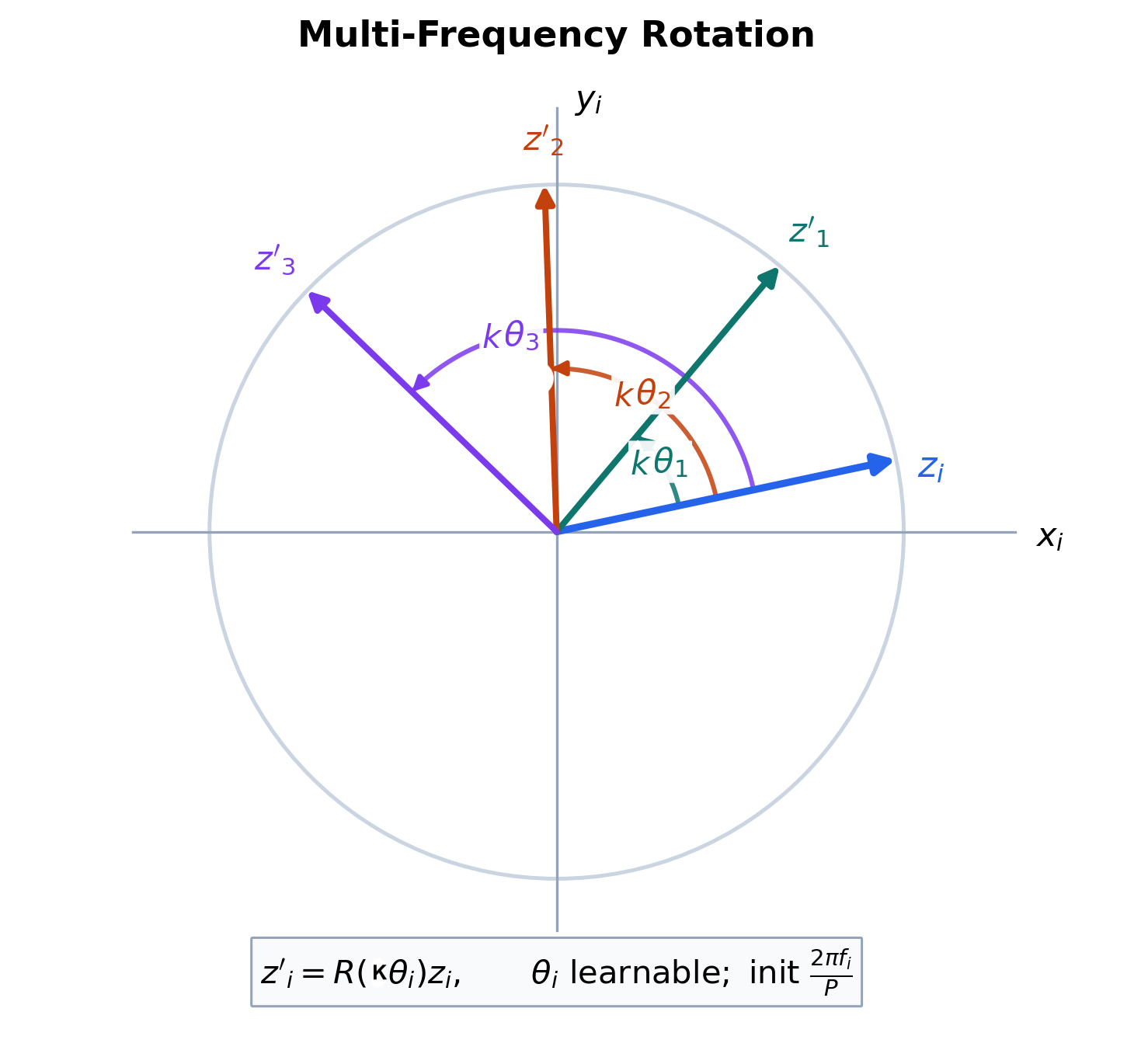}
    \caption{Working of multi-frequency rotation, each pair of $z$ is rotated by different angles.}
    \label{fig:mfr_demo}
\end{figure}

\subsection{Variance-Invariance-Covariance Regularization (VICReg)}
\label{sec:vicreg}

The model optimizes the objective $\mathcal{L} = \lambda \mathcal{L}_{\text{inv}} + \mu \mathcal{L}_{\text{var}} + \nu \mathcal{L}_{\text{cov}}$ to prevent latent collapse. The L2 invariance loss ($\mathcal{L}_{\text{inv}}$) aligns the predicted and target latents, i.e., the standard JEPA loss. The variance loss ($\mathcal{L}_{\text{var}}$) is a hinge loss that incentivizes the standard deviation of each latent dimension across the batch to be above a certain threshold (1 in this work). The covariance loss ($\mathcal{L}_{\text{cov}}$) minimizes the covariance between each pair of latent dimensions across the batch. In this way, VICReg prevents representations of different samples in a batch (including those from different classes) from collapsing to the same trivial invariant features, thereby avoiding latent collapse.

For VICReg to work, a relatively large batch size is required. For all experiments, we use a batch size of 128. We use the default parameter values from the original VICReg paper: $\lambda = 25, \mu = 25, \nu = 1$.

\subsection{SFR degrades in learnable rotation angles}
\label{sec: sfr_degrades}
\begin{figure}[t]
\begin{center}
\includegraphics[width=0.5\linewidth]{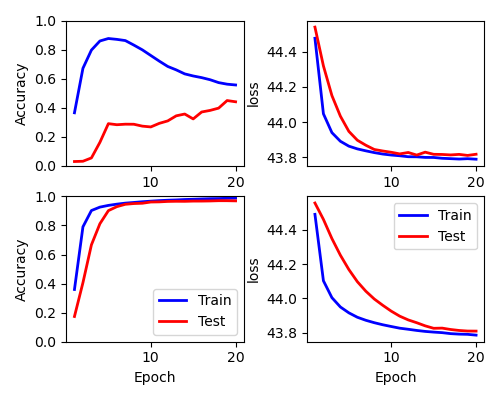}
\end{center}
   \caption{Learning curves show the accuracy (left) and loss (right) during training (blue) and test(red) for SFR (top) and MFR (bottom)}
\label{fig:supple_lc}

\end{figure}

\begin{figure}[t]
\begin{center}
\includegraphics[width=0.8\linewidth]{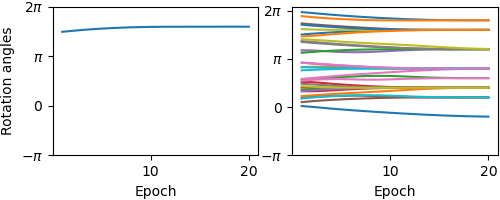}
\end{center}
   \caption{Learning curves show the learned angle(s) in SFR (left) and MFR(right) during training}
\label{fig:supple_lc_angle}

\end{figure}

With a different initial condition of $\theta$s ($\in [0, 2\pi]$), MFR still learns robustly with both ~97\% accuracy in both training and testing, while SFR significantly degrades its performance in both settings. (Fig. \ref{fig:supple_lc}). Correspondingly, $\theta$ in SFR converges to a higher-order harmonics $8\pi/5$ instead of the base frequency $\pi/5$ (Fig. \ref{fig:supple_lc_angle}, left), which breaks the compositional modulo-10 structure in the latent space needed for arithmetic operation. MFR on the other hand, shows the flexibility in learning with multiple independent frequencies (Fig. \ref{fig:supple_lc_angle}, right).    

\subsection{Geometric Structure of Latent Space }
\label{sec: geometry_latent}

The $+1$ trajectory in Fig. \ref{fig:traj_plus1_sup} demonstrates the model's recurrent stability. Despite relying on purely its own predictions for 10 sequential gaps, the model accurately visits every class prototype with minimal drift upon returning to the start state. This demonstrates the block rotation operator applies rigid transformations without compounding error.

\subsection{Multi-Frequency Rotation - PCA Plots}
 In the MFR PCA visualization, digit pairs separated by five modular steps, such as $0/5$, $1/6$, $2/7$, $3/8$, and $4/9$, appear close in the two-dimensional projection. This does not necessarily indicate latent collapse. Since MFR uses multiple rotation frequencies, the $+5$ transformation affects different latent blocks differently: some frequency components make these pairs antipodal, while others make them similar. PCA shows only the dominant two directions of variance, so it can group these half-cycle pairs together even when they remain separable in the full latent space. Consistent with this, the MFR heatmap in Fig.~\ref{fig:heatmap_comparison} shows only moderate similarity for these pairs rather than near-perfect correlation, indicating that class information is still distributed across the remaining latent dimensions.

\begin{figure}[h]
  \centering
  \includegraphics[width=0.5\linewidth]{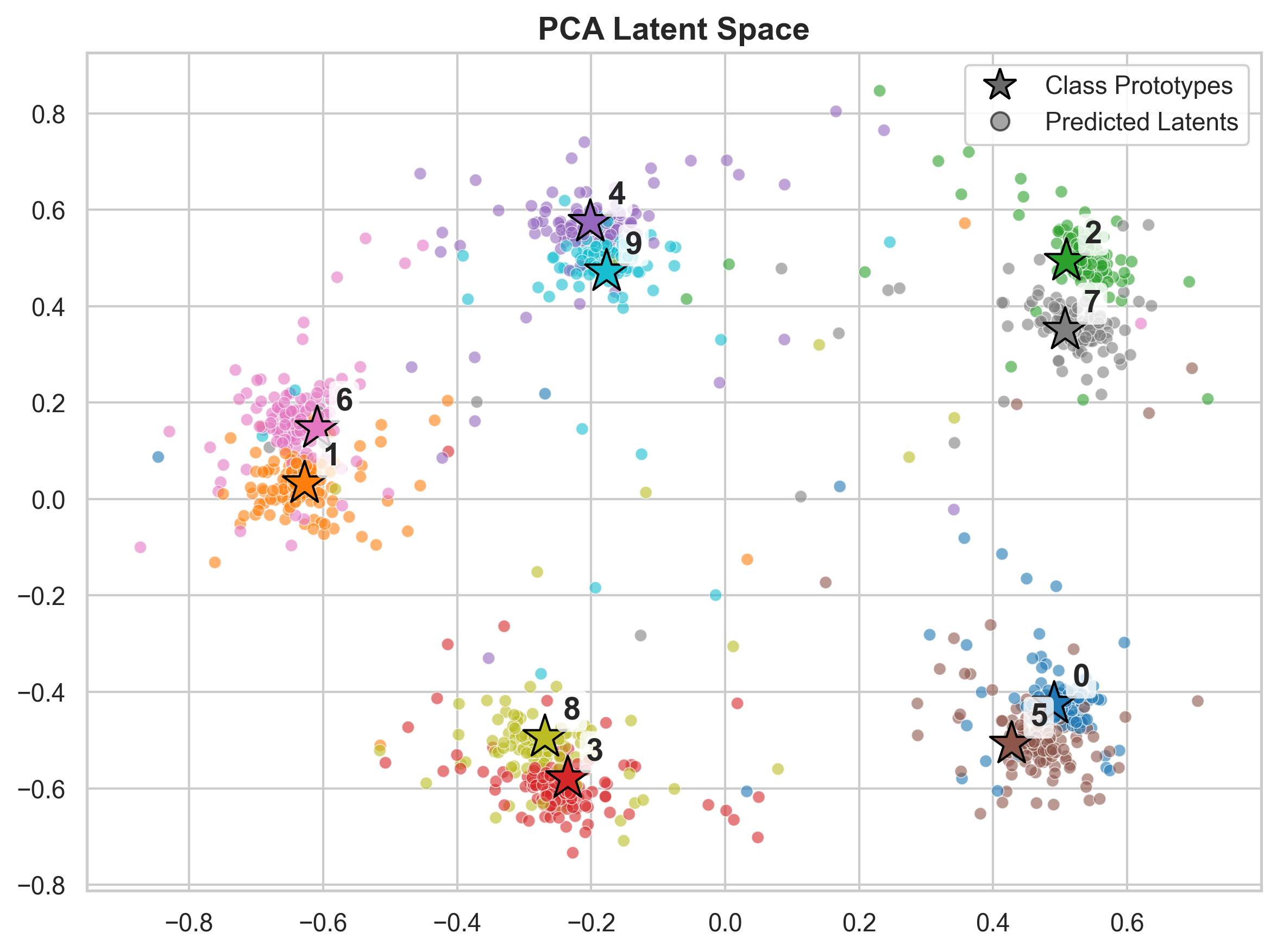}
  \caption{Latent space PCA of MFR block-rotation predictor. By explicitly enforcing rotational dynamics, the model successfully disentangles the latent space into a circular manifold representing the modulo-10 arithmetic structure.}
  \label{fig:blockrot_pca_MFR}
\end{figure}

\begin{figure*}[t]
  \centering
  \begin{subfigure}{0.48\textwidth}
    \includegraphics[width=\linewidth]{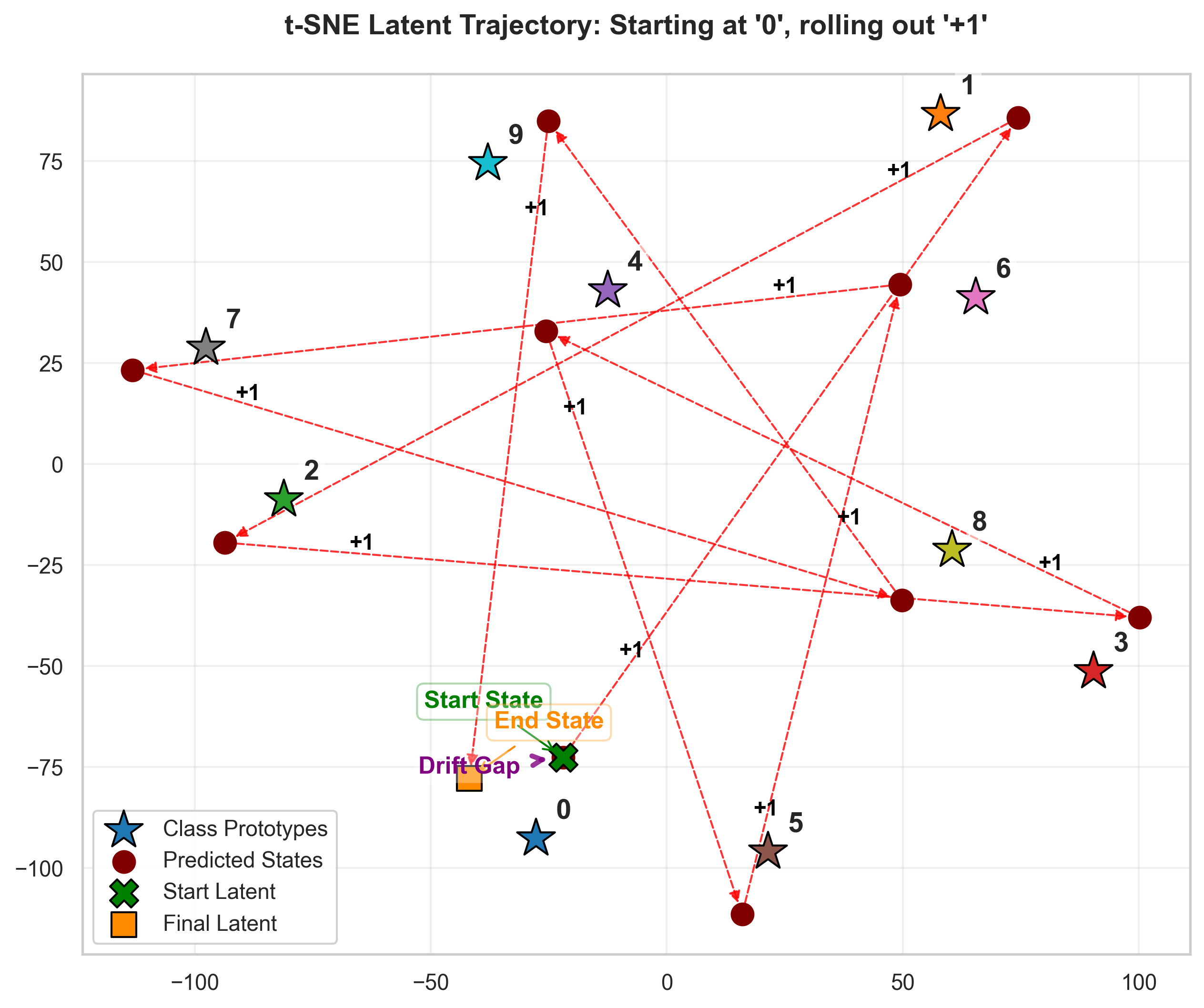}
    \caption{Rollout for $+1$ (In-Distribution)}
    \label{fig:traj_plus1_sup}
  \end{subfigure}
  \hfill
  \begin{subfigure}{0.48\textwidth}
    \includegraphics[width=\linewidth]{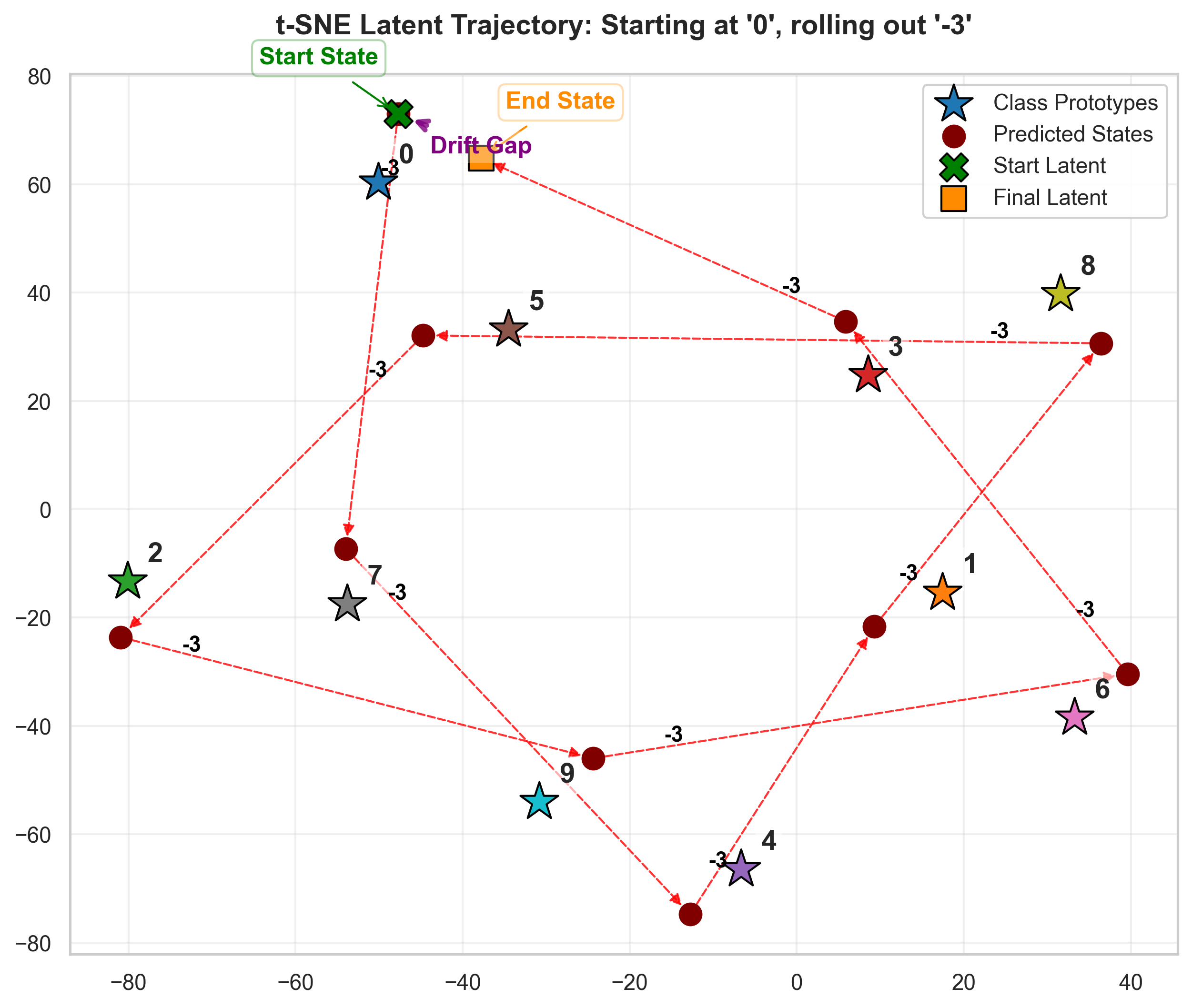}
    \caption{Rollout for $-3$ (Zero-Shot)}
    \label{fig:traj_minus3_sup}
  \end{subfigure}
  \caption{t-SNE visualization of multi-step latent trajectories for the Block Rotation predictor (MFR). Starting from an initial state of $0$, the model demonstrates high stability over 10 sequential steps for trained operations (a) and successfully composes unseen operations with a minimal drift gap (b).}
  \label{fig:trajectories_sup}
\end{figure*}

\subsection{EMNIST Letters Implementation Details}
We  (1) invert EMNIST's default $90^\circ$ rotation and horizontal flip in the dataloader, (2) shift the 1-indexed EMNIST labels to 0-indexing for consistency with our MNIST setup, and (3) recompute image normalization statistics for EMNIST.

\end{document}